\documentclass{article}

    \usepackage[preprint]{neurips_2025}

\usepackage{tikz}
\usetikzlibrary{arrows.meta}
\usepackage{pgfplots}

\usepackage[utf8]{inputenc} %
\usepackage[T1]{fontenc}    %
\usepackage{hyperref}       %
\usepackage{url}            %
\usepackage{booktabs}       %
\usepackage{amsfonts}       %
\usepackage{nicefrac}       %
\usepackage{microtype}      %
\usepackage{xcolor}         %

\usepackage{graphicx}
\usepackage{xspace}

\usepackage{algorithm}
\usepackage{algpseudocode}

\usepackage{amsmath}
\usepackage{amssymb}
\usepackage{mathtools}
\usepackage{amsthm}

\usepackage{multirow}
\usepackage{wrapfig} 
\usepackage{colortbl}

\usepackage{amsmath,amsfonts,bm}

\def\eqref#1{equation~\ref{#1}}

\def\1{\bm{1}}

\def\rmK{{\mathbf{K}}}

\def\rmM{{\mathbf{M}}}

\def\rmO{{\mathbf{O}}}

\def\rmQ{{\mathbf{Q}}}

\def\rmS{{\mathbf{S}}}

\def\rmV{{\mathbf{V}}}

\def\rmX{{\mathbf{X}}}
\def\rmY{{\mathbf{Y}}}

\def\vk{{\bm{k}}}

\def\vo{{\bm{o}}}

\def\vq{{\bm{q}}}

\def\vv{{\bm{v}}}

\def\vx{{\bm{x}}}
\def\vy{{\bm{y}}}
\def\vz{{\bm{z}}}

\def\mW{{\bm{W}}}

\DeclareMathAlphabet{\mathsfit}{\encodingdefault}{\sfdefault}{m}{sl}
\SetMathAlphabet{\mathsfit}{bold}{\encodingdefault}{\sfdefault}{bx}{n}

\newcommand{\R}{\mathbb{R}}

\newcommand{\softmax}{\mathrm{softmax}}

\providecommand{\rms}{\text{RMS}}

\algblock{ParFor}{EndParFor}
\algnewcommand\algorithmicparfor{\textbf{parfor}}
\algnewcommand\algorithmicpardo{\textbf{do}}
\algnewcommand\algorithmicendparfor{\textbf{end\ parfor}}
\algrenewtext{ParFor}[1]{\algorithmicparfor\ #1\ \algorithmicpardo}
\algrenewtext{EndParFor}{\algorithmicendparfor}

\newcommand{\sysname}{\textsc{RAttention}\xspace}

\title{\sysname: Towards the Minimal Sliding Window Size in Local-Global Attention Models}

\author{%
Bailin Wang \hspace{2mm} Chang Lan \hspace{2mm} Chong Wang \hspace{2mm} Ruoming Pang \\
Apple \\
\texttt{
\{bwang47,c\_lan,mr.chongwang,rpang\}@apple.com}
}

\begin{document}

\maketitle

\begin{abstract}

Local-global attention models~\cite{characterAI2024,gemmateam2024gemma2} have recently emerged as compelling alternatives to standard Transformers, promising improvements in both training and inference efficiency. However, the crucial choice of window size presents a Pareto tradeoff: larger windows maintain performance akin to full attention but offer minimal efficiency gains in short-context scenarios, while smaller windows can lead to performance degradation. Current models, such as Gemma2 and Mistral, adopt conservative window sizes (e.g., 4096 out of an 8192 pretraining length) to preserve performance. This work investigates strategies to shift this Pareto frontier, enabling local-global models to achieve efficiency gains even in short-context regimes.
Our core motivation is to address the intrinsic limitation of local attention—its complete disregard for tokens outside the defined window. We explore \sysname, a variant of local attention integrated with a specialized linear attention mechanism designed to capture information from these out-of-window tokens. Pretraining experiments at the 3B and 12B scales demonstrate that \sysname achieves a superior Pareto tradeoff between performance and efficiency. As a sweetspot, \sysname with a window size of just 512 consistently matches the performance of full-attention models across diverse settings.
Furthermore, the recurrent nature inherent in the linear attention component of \sysname contributes to enhanced long-context performance, as validated on the RULER benchmark~\cite{hsieh2024ruler}. Crucially, these improvements do not compromise training efficiency; thanks to a specialized kernel implementation and the reduced window size, \sysname maintains training speeds comparable to existing state-of-the-art approaches.~\footnote{Our JAX implementation of \sysname using Pallas kernels are open-sourced at \url{https://github.com/apple/axlearn/tree/main/axlearn/common/rattention}.}
\end{abstract}

\section{Introduction}

Improving the efficiency of standard Transformers has been a central focus of architectural design. Sliding Window Attention (SWA)~\cite{child2019swa}, a natural variant of standard attention, has been widely adopted to reduce both memory transfer and computational costs, as demonstrated in prominent models like Mistral~\cite{jiang2023mistral7b} and Gemma~\cite{gemmateam2024gemma2}. Its constant memory consumption during decoding is particularly appealing for decoding time which is bounded by memory transfers rather than computation. However, SWA presents an inherent Pareto tradeoff between model capacity and efficiency: increasing the window size improves performance but diminishes efficiency gains. Recent models have often adopted conservative window sizes; for instance, both Mistral and Gemma utilize a 4096-token window out of a 8192-token pretraining context. Consequently, the efficiency benefits of SWA only become substantial for relatively long sequences. To illustrate, a 12B parameter local-global attention model we tested, using a 4K window size, apparently offers no KV cache savings for $\leq$ 4K context tasks. However, reducing the window size to 1k could yield approximately 56\% KV cache savings at 4K length. In this work, we investigate whether this Pareto frontier can be shifted -- achieving comparable performance with a significantly smaller window size.

\begin{figure}[t]
\centering
\includegraphics[width=0.9\textwidth]{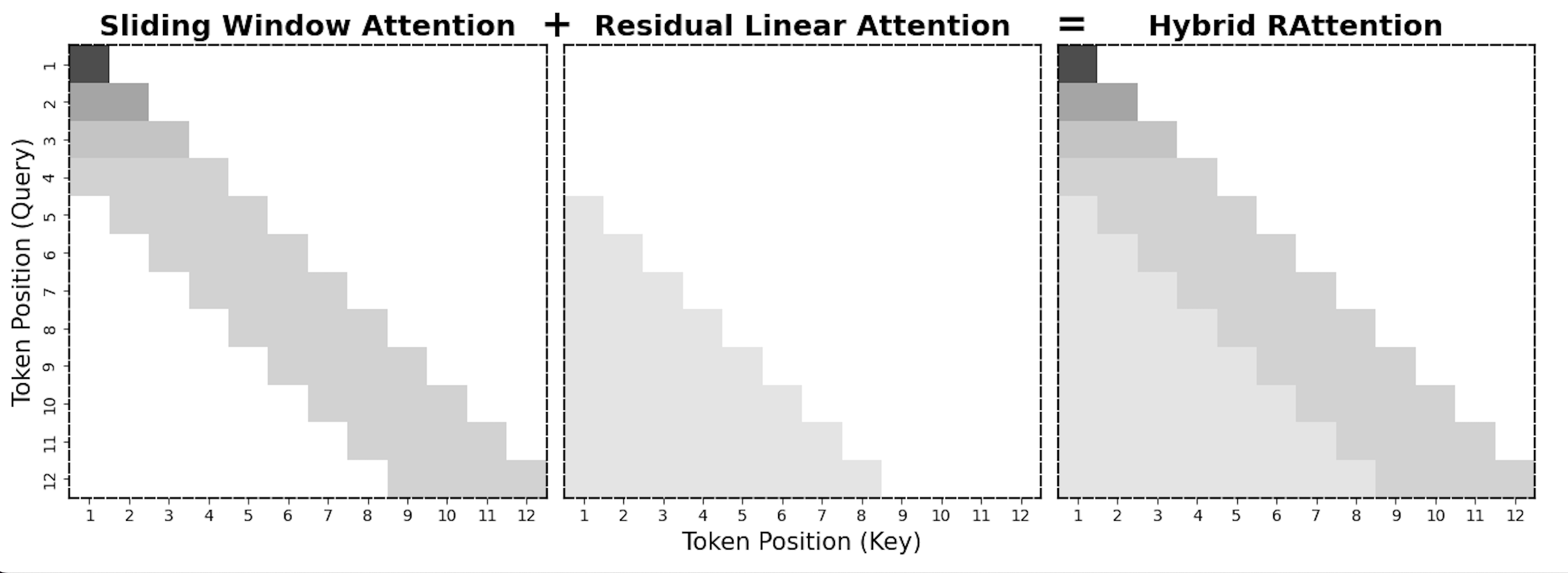}
\vspace{-3mm}
\caption{\sysname combines Sliding Window Attention (SWA) for local context with Residual Linear Attention (RLA) to gather information from out-of-window tokens.  Apart from 4 in-window tokens, RLA compresses the information of token 1 for query token 5; token 1,2 for query token 6.}
\label{fig:intro-mask}
\vspace{-6mm}
\end{figure}

Our primary hypothesis is that the performance degradation observed when reducing the SWA window size stems from an intrinsic limitation of local attention: simply ensuring that \textit{number\_of\_layers $\times$ window\_size $\geq$ context\_length} is often an inadequate heuristic for determining the optimal window size. To address this, we explore the integration of a Residual Linear Attention (RLA) module. RLA, in our design, is a specialized linear attention model~\cite{katharopoulos2020linear} aimed at capturing information from "residual tokens" -- those lying beyond the immediate SWA window. This RLA module inherently enhances the capabilities of SWA. The recurrent nature of RLA offers two significant advantages: first, it preserves the constant-memory property crucial for efficient decoding with SWA; second, it lessens the over-reliance on positional embeddings for length extrapolation, leading to markedly better zero-shot length generalization compared to SWA alone. As depicted in Figure~\ref{fig:intro-mask}, the resulting hybrid architecture, which we term \sysname (Sliding Window \textit{Attention} with \textit{R}esidual Linear Attention), effectively captures information from all tokens within the context.

The key contribution of this work is demonstrating that \sysname can serve as a drop-in replacement for standard local attention mechanisms, positioning local-global models with \sysname as practical alternatives to full-attention Transformers. First, we show that \sysname models~\footnote{Throughout this paper, "SWA models" and "\sysname models" refer to hybrid architectures that combine global attention with either SWA or \sysname, respectively, unless specified otherwise.} can match, and in some cases outperform, SWA models that use much larger sliding window sizes, while offering superior inference efficiency. Second, we demonstrate that this improved efficiency does not compromise training speed; with our dedicated kernel implementations and the reduced window size, \sysname models maintain training efficiency comparable to SWA models with larger windows. While the concept of combining SWA with linear attention has been explored (e.g., in \citep{arora2025based,zhang2025lolcats}, where a tiny SWA window complements linear attention), those approaches did not match the performance of full-attention models. We show that local-global models incorporating \sysname can be reliable alternatives to standard full-attention Transformers.

We conduct extensive experiments comparing \sysname with SWA within a local-global hybrid attention framework. Our results demonstrate that \sysname consistently matches or surpasses the performance of SWA-based models while requiring significantly less memory due to smaller window sizes. Furthermore, \sysname yields substantial performance improvements on long-context reasoning tasks, as measured by the RULER benchmark~\cite{hsieh2024ruler}. On the training efficiency side, we develop dedicated kernels for \sysname so that the training efficiency is not compromised compared with full-attention models.

\section{Background}

We first briefly introduce standard attention and linear attention mechanisms. For notation, we use uppercase letters for matrices, lowercase letters for row vectors.

\subsection{Global Attention and Local Attention}

In standard Transformers, an input sequence $\rmX \in \R^{L \times d}$ (here $L$ is the length and $d$ is the hidden dimension) in the attention layer is processed through,
\begin{align*}
\vq_t, \ \vk_t, \ \vv_t &= \vx_t \mW_Q, \  \vx_t \mW_K, \ \vx_t \mW_V, \\ \vo_t &= \frac{\sum_{i=1}^{t} \exp(\vq_t  \vk_i^\intercal)\vv_i}{\sum_{i =1} ^{t} \exp(\vq_t  \vk_i^\intercal)},
\end{align*}
which  computes the query ($\vq_t$), key ($\vk_t$), and value ($\vv_t$) vectors given the current token's representation $\vx_t \in \R^{1 \times d}$. Then attention is performed over the all the previous keys $\{\vk_1, \dots, \vk_t\}$ and values  $\{\vv_1, \dots, \vv_t\}$. During inference, as the $t$ grow, the set of keys and values (i.e., KV cache) also grows linearly, leading to heavy memory consumption if $t$ is large. Local attention (i.e., sliding window attention)~\cite{child2019swa} is then introduced to achieve constant-memory consumption during inference. 

The basic idea of SWA is to limit the attention to only recent $w$ tokens through,
\begin{align*}
\vo^{\text{\scriptsize{swa}}}_t &= \frac{\sum_{i=\max(1, t - w)}^{t} \exp(\vq_t  \vk_i^\intercal)\vv_i}{\sum_{i =\max(1, t - w)} ^{t} \exp(\vq_t  \vk_i^\intercal)}.
\end{align*}
As a result, at most $w+1$ tokens (including the current token) need to be considered as KV cache during inference. The choice of $w$ is based on the Pareto tradeoff between downstream performance and efficiency, and we aim to shift the Pareto frontier in this work.

\subsection{Linear Attention}

Linear attention mechanisms~\cite{katharopoulos2020linear}  replace $\exp(\vq_t \vk_i^\intercal)$ with a kernel $k(\vx, \vy)$ with an associated feature map $\phi$ (i.e., $k(\vx, \vy) = \langle\phi(\vx), \phi(\vy)\rangle$).  This simplifies the calculation of $\vo_t$ since we have
\begin{align*}
 \vo^{\text{\scriptsize{la}}}_t &= \frac{\sum_{i=1}^{ t}\phi(\vq_t)\phi(\vk_i)^\intercal \vv_i}{\sum_{i=1}^{t} \phi(\vq_t)\phi(\vk_i)^\intercal  }  
 = \frac{\phi(\vq_t)   \sum_{i=1}^{t}\phi(\vk_i)^\intercal \vv_i}{\phi(\vq_t) \sum_{i=1}^{t}\phi(\vk_i)^\intercal}.
\end{align*}
Letting $\rmS_t=\sum_{i=1}^{t}\phi(\vk_i)^\intercal \vv_i$ and $\vz_t=\sum_{i=1}^{t}\phi(\vk_i)^\intercal$ where $\rmS_t \in \mathbb{R}^{d'\times d}, \vz_t \in \mathbb{R}^{d'\times 1}$, we can rewrite the above as an RNN,
\begin{align*}
\rmS_t = \rmS_{t-1} &+ \phi(\vk_t)^\intercal \vv_t, \hspace{1mm} \vz_t = \vz_{t-1} + \phi(\vk_t)^\intercal, \hspace{1mm} \vo^{\text{\scriptsize{la}}}_t = \frac{\phi(\vq_t) \rmS_t}{ \phi(\vq_t) \vz_t}.
\end{align*}
where $d'$ denotes the output size of feature function $\phi$, initial state $\rmS_{0} = \bm 0$.
Recent work has found that the normalization terms can be replaced with a simpler RMSNorm on the output. 
\begin{align}
& \rmS_t = \rmS_{t-1} + \phi(\vk_t)^\intercal \vv_t, \quad \vo^{\text{\scriptsize{la}}}_t =  \phi(\vq_t) \rmS_t .
\label{eq:simple_linear_attention}
\end{align}
Eq.~\ref{eq:simple_linear_attention} makes it clear that  a linear attention layer is essentially a linear recurrent layer with matrix-valued hidden states $\rmS_t$ that is updated via the outer-product $\phi(\vk_t)^\intercal \vv_t$.

\paragraph{Chunkwise Parallel Form}

The recurrent form in Eq.\ref{eq:simple_linear_attention} highlights the inherently sequential nature of linear attention, which makes it advantageous during the decoding stage. During training or prefilling, we rely on an equivalent chunkwise parallel formulation to compute the output efficiently\cite{yang2024gla}.

Let the tilded input $\rmQ_{[i]} := \rmQ_{iC+1:(i+1)C+1} \in \mathbb{R}^{C \times d}$ represent the query vectors for the $i$-th chunk, and define $\rmK_{[i]}$, $\rmV_{[i]}$, and $\rmO_{[i]}$ similarly. The output for the chunk can be computed as:
\begin{align}    
\rmS_{[i+1]} = \rmS_{[i]} + \tilde \rmK^\intercal_{[i]}\rmV_{[i]} \quad \hspace{1mm} \in \mathbb{R}^{d'\times d} \\
\rmO_{[i]} = \underbrace{\rmQ_{[i]}\rmS_{[i-1]}}_{\rmO^\text{inter}_{[i]}} + \underbrace{\big(((\tilde  \rmQ_{[i]}) \tilde \rmK_{[i]}^{\intercal})\odot\rmM\big)\rmV_{[i]}}_{\rmO^{\text{intra}}_{[i]}},
\label{eq:la_chunking}
\end{align}
Here, $C$ denotes the chunk (tiling) size, and $\tilde\rmQ = \phi(\rmQ)$, $\tilde\rmK = \phi(\rmK)$. The key insight is that, given the chunk-level state $\rmS_{[i]}$, the output within each chunk can be computed in parallel using matrix multiplications -- an operation that modern accelerators are highly optimized for.

Compared to global attention, linear attention requires maintaining only a fixed-size state S during inference, similar to local attention. Specifically, its cache is equivalent to using a window size of $\frac{d'\times d}{2 \times d}$ in SWA~\footnote{The factor of 2 in the denominator accounts for both key and value storage. d = 128 in a multi-head setting.}. While this leads to significant inference efficiency, pure linear attention models still fall short of Transformers, particularly in recall-intensive tasks~\cite{akyürek2024icl}.

\section{Method}

\subsection{Residual Linear Attention (RLA)}

To address the limitation of SWA, we use a specialized linear attention (RLA) that processes out-of-window tokens, as illustrated in the middle figure in  Figure~\ref{fig:intro-mask}.
Concretely, the recurrence of RLA is as follwows,
\begin{align}
& \rmS_t = \rmS_{t-1} + \phi(\vk_t)^\intercal \vv_t, \quad \vo^{\text{\scriptsize{rla}}}_t =  \phi(\vq_t) \rmS_{t - w - 1}  .
\label{eq:residual_linear_attention}
\end{align}
where instead of reading out from $\rmS_t$ as in Eq~\ref{eq:simple_linear_attention}, $\vo_{t}$ is obtained via reading out from $\rmS_{t-w-1}$ -- the hidden states that captures the contextual information ends at token $(t - w - 1)$. 
To integrate with SWA, we combine he outputs from RLA and SWA, as demonstrated in Figure~\ref{fig:intro-mask}.

\paragraph{Parameterization} Employing separate parameters (i.e., projections for query/key/value/output) for RLA and SWA would double 
 the parameter size of token-mixing layers. In this work, we find that simply \textit{shares all parameters is sufficient}, provided that an appropriate feature map is used. As we will show in the experiments later, the softmax-based feature map~\cite{zhang2025lolcats} yields the best performance, whereas the identity feature map, commonly used in pure linear attention models, is less effective.

\paragraph{Group-Query Variant} We extend RLA to support Group-Query Attention (GQA), a popular variant of Multi-Head Attention that reduces KV memory usage. Specifically, we share key and value vectors within each query group, maintaining efficiency while preserving model quality. When coupled with SWA, RLA will share the same grouping structures as SWA as well.

\subsection{Hybrid \sysname Models}

We combine the results from RLA and SWA as follows:
\begin{align*}
    \vo_t = \rms ( \vo^{\text{\scriptsize{swa}}}_t ) + \rms ( \vo^{\text{\scriptsize{rla}}}_t ) 
\end{align*}
where two separate RMS norms are used upon the output from SWA and RLA respectively.
In the group-query head variant, different heads will use separate parameters for RMS, which is a common strategy used in linear attention models~\cite{sun2023retentive,yang2024gla,dao2024mamba2}.
Collectively, we call the resulting hybrid attention module \sysname.

\paragraph{Parameter Efficiency}

Our design introduces no additional parameters over standard SWA modules, as \sysname\ fully reuses the existing query/key/value projections. In contrast, recent work~\cite{dong2024hymba} that combines state-space models (SSM) with SWA requires separate parameter sets for SSM and SWA. This results in a significantly larger portion of the model’s parameters being allocated to token-mixing layers, which potentially limit the capacity of the feedforward layers when scaling up.

\paragraph{Kernel Design}

\usetikzlibrary{arrows.meta,      %
                positioning,       %
                shadows}           %

\usetikzlibrary{shapes.multipart,  %
                positioning}       %

\usetikzlibrary{shapes,            %
                arrows,            %
                arrows.meta,       %
                fit,               %
                backgrounds,       %
                positioning,       %
                calc}              %

\tikzstyle{process} = [rectangle, 
                      minimum width=2cm, 
                      minimum height=1cm, 
                      text centered, 
                      text width=2cm,  
                      draw=black, 
                      fill=orange!10]

\tikzstyle{arrow} = [thick,->,>=stealth]  %

\tikzset{
    font=\sffamily,  %
    Q_BLOCK/.style={
        draw,
        align=center,
        draw=orange!60,
        fill=orange!10,
        rectangle,
        text width=0.9cm,
        minimum height=0.6cm,
    },
    KV_BLOCK/.style={
        draw,
        align=center,
        draw=orange!60,
        fill=orange!10,
        rectangle split, 
        rectangle split horizontal,  %
        rectangle split parts=2,     %
    },
    O_BLOCK/.style={
        draw,
        align=center,
        draw=purple!50,
        fill=purple!15,
        rectangle,
        text width=0.92cm,
        minimum height=0.75cm,
    }
}

\begin{figure}[t]  %
\centering  
\resizebox{.8\columnwidth}{!}{  %
    \begin{tikzpicture}[%
        every path/.style={thick},  %
        dcs/.style = {double copy shadow,  %
                     shadow xshift=4pt, 
                     shadow yshift=-4pt}
      ]
    \node [rectangle, minimum width=.5cm, minimum height=0.8cm, 
           text centered, text width=1cm, 
           draw=black, fill=orange!10] (s_prev) at (-3.0, -1.5) {$\mathbf{S}_{[i-1]}$};

    \node [rectangle, minimum width=.5cm, minimum height=0.8cm, 
           text centered, text width=1cm,  
           draw=black, fill=teal!25] (s_i) at (0.8, -1.5) {$\rmS_{[i]}$};
           
    \node (dots_s) at (2.45, -2) {$\ldots$};
           
    \node [rectangle, minimum width=.5cm, minimum height=0.8cm, 
           text centered, text width=1.1cm,  
           draw=black, fill=teal!25] (s_im1) at (4.3, -1.5) {$\rmS_{[\scriptscriptstyle i+m-1]}$
};
           
    \node [rectangle, minimum width=.5cm, minimum height=0.8cm, 
           text centered, text width=1cm,  
           draw=black, fill=purple!15] (s_im) at (8.3, -1.5) {$\rmS_{[i+m]}$};

    \node[O_BLOCK] (o_inter_i) at (-3.5, 0) {$\rmO^{\text{inter}}_{[i]}$};
    \node[O_BLOCK] (o_intra_i) at (0.8, -0.3) {$\rmO^{\text{intra}}_{[i]}$};

    \node[O_BLOCK] (o_inter_im) at (4, 0) {$\rmO^{\text{inter}}_{[i+m]}$};
    \node[O_BLOCK] (o_intra_im) at (8.3, -0.3) {$\rmO^{\text{intra}}_{[i+m]}$};

    \node[Q_BLOCK] (q_block_i) at (-1.0,-2.5) {$\rmQ_{[i]}$};
    
    \node[KV_BLOCK] (kv_block_i) at (-0.5,-3.5) {
        \nodepart{one} $\rmK_{[i]}$ 
        \nodepart{two} $\rmV_{[i]}$
    };

    \node[Q_BLOCK] (q_block_im) at (6,-2.5) {$\rmQ_{[\scriptscriptstyle i+m]}$};
    
    \node[KV_BLOCK] (kv_block_im) at (7.2,-3.5) {
        \nodepart{one} $\rmK_{[\scriptscriptstyle i+m]}$ 
        \nodepart{two} $\rmV_{[\scriptscriptstyle i+m]}$
    };

    \node [rectangle, text centered, text width=.5cm, text height=.5cm, 
           draw=black, fill=orange!10, 
           label=right:{Load from HBM}] (node1) at (-4.0, -5){};
           
    \node [rectangle, text centered, text width=.5cm, text height=.5cm,
           draw=black, fill=purple!15, 
           label=right:Store to HBM] (node2) at (0.5, -5){};
    
    \node [rectangle, text centered, text width=.5cm, text height=.5cm, 
           draw=black, fill=teal!25, 
           label=right:On-chip] (node3) at (4.2, -5){};

    \node [rectangle, text centered, text width=1.5cm, text height=0.3cm,
       draw=none, label={right:\phantom{xx}}] (arrow_legend) at (-3.5, -4.2) {};
        \draw[arrow, dashed] (6.8,-5) -- (7.6,-5);
        \node[text width=2cm] at (8.8, -5) {Sequential};

    \draw [arrow, dashed] (s_prev) -- (s_i);
    \draw [arrow, dashed] (s_im1) -- (s_im);

    \draw [arrow, dashed] ($(s_prev)-(2.0,0)$) -- (s_prev);
    \draw [arrow, dashed] ($(s_im1)-(1.5,0)$) -- (s_im1);
    \draw [arrow, dashed] (s_i) -- ($(s_i)+(1.2,0)$);
    \draw [arrow, dashed] (s_im) -- ($(s_im)+(1.8,0)$);

\tikzstyle{container1} = [draw=black, rectangle, inner sep=0.5cm, fill=black!2];
\tikzstyle{container2} = [draw=blue!70, thick, dashed, inner sep=0.3cm, fill=blue!15, opacity=0.5];
\tikzstyle{container3} = [draw=green!70, thick, dashed, rectangle, inner sep=0.3cm, fill=green!15, opacity=0.5];

\begin{scope}[on background layer]
    \node [container1] 
          (contain1) [fit= {(s_prev) (s_i) (s_im1) (q_block_i) (o_inter_i) (o_intra_i) 
                           (s_im) (q_block_im) (kv_block_i) (kv_block_im) (dots_s) 
                           (o_inter_im) (o_intra_im) ($(o_intra_im)+(1cm,1cm)$) },
                       ] {};

    \node [container2] 
          (contain2a) [fit= {(s_prev) (q_block_i) (o_inter_i)},
          label={[font=\small, anchor=south west]south west:\textit{inter-chunk}}] {};
    
    \node [container2] 
          (contain2b) [fit= {(s_im1) (q_block_im) (o_inter_im)}, label={[font=\small, anchor=south west]south west:\textit{inter-chunk}}] {};

    \node [container3] 
          (contain3a) [fit= {(q_block_i) (kv_block_i) ($(kv_block_i)+(0,-0.4cm)$) (o_intra_i) (s_i)},
          label={[font=\small, anchor=south east]south east:\textit{intra-chunk}}] {};
    
    \node [container3] 
          (contain3b) [fit= {(q_block_im) (kv_block_im) ($(kv_block_im)+(0,-0.4cm)$) (o_intra_im) (s_im)}, label={[font=\small, anchor=south east]south east:\textit{intra-chunk}}] {};
\end{scope}

\end{tikzpicture}   
}
    \vspace{-2mm}
\caption{Interleaved state-saving pattern used in our training kernels. For every $m$ chunks, only the state of the last chunk ($\rmS_{[m]}$) is stored in HBM. The intermediate chunk states are recomputed on-chip as needed, using the most recent stored state $\rmS_{[i-1]}$ from the previous group of $m$ chunks. This approach make it more flexible to balances memory I/O cost and matmul 
 computation.}
    \label{fig:chunking}
\vspace{-4mm}
\end{figure}

Our linear attention kernels incorporate two key optimizations—fused operations and flexible state saving. To reduce memory I/O cost, we first fuse the computation of the feature map \( \phi(\vk_t, \vv_t) \) directly within the kernel, eliminating the need to store intermediate values in HBM and transfer them to VMEM, thereby lowering memory overhead.

Beyond reducing memory I/O, another critical factor for efficiency is maximizing the overlap between memory access and computation. In chunk-wise training, there is an inherent tradeoff in how much intermediate state ($\rmS_{[i]}$) to store: storing more leads to faster backward passes (which depend on these states) but comes at a higher memory cost. Typically, chunk size is tuned according to two strategies -- either storing all intermediate states or recomputing them during the backward pass. In this work, we introduce a more flexible state saving scheme. As illustrated in Figure~\ref{fig:chunking}, we store states every $m$ chunks while recomputing the states for the intermediate chunks. This hybrid approach allows compilers (e.g., using Triton or Pallas) to more effectively schedule operations and overlap memory I/O with computation. By enumerative search over $m$ and chunk size, the best configuration can achieve around 15\% speedup compared with $m=1$ and a typical chunk size 256.

\subsection{Local-Global Attention Models}

We evaluate \sysname\ within a local-global attention framework that stacks multiple blocks in a repeating pattern of \texttt{[local, local, local, global]}.~\footnote{The ration between local and global attention is another axis to consider when exploring the tradeoff between downstream performance and efficiency of local-global models. We leave this direction for future work.}
The computation in each block follows:
\begin{align*}
&\rmY^{(l)} = \operatorname{Attention}^{(l)}(\rms(\rmX^{(l)})) + \rmX^{(l)} \
&\rmX^{(l+1)} = \operatorname{SwiGLU}(\rms(\rmY^{(l)})) + \rmX^{(l)},
\end{align*}
where $l$ indexes the layer starting from $1$. The attention module $\operatorname{Attention}^{(l)}$ is instantiated as \sysname when $l \mod 4 \neq 0$, and as standard global attention otherwise. Following the LLaMA architecture~\cite{touvron2023llama}, we adopt pre-norm and SwiGLU feed-forward layers. 

While the combination of linear attention and sliding window attention was previously investigated by~\cite{arora2025based}, their models, despite outperforming other purely linear architectures, did not achieve performance parity with full-attention Transformers, even at the 1B scale. \sysname local-global models, in contrast, successfully bridges this performance gap. %
\section{Experiment}

\subsection{Main Results}

The central question we address in our experiments is: \textit{what is the Pareto trade-off between downstream performance and efficiency (i.e., sliding window size) when replacing SWA with \sysname}? Our evaluation consists of two stages: 
\begin{enumerate}
    \item  we establish the trade-off curve between performance and window size through extensive experiments at the 3B scale. From this curve, we identify the minimal window size range for \sysname that achieves performance comparable to full attention;
    \item we further validate these window sizes in settings where models are scaled by in three axes: the number of tokens, the model size or the pretraining context length.
\end{enumerate}
The short answer to the question is that \sysname with a window size of $\geq 512$ can reliably match (or exceed) the performance of full attention across different settings. 

\paragraph{Pretraining Setup}
All the models are implemented in Jax~\cite{jax2018github} and trained on v6e Cloud TPU clusters. We use 512/1024 chips provided as 2/4 $\times$ 256 chip slices to train 3B/12B models, respectively. Data parallelism along with activation recomputation is used for distributed training. We use a variant of RMSProp~\cite{rmsprop} with momentum as the optimizer.
We use our internal web-crawled data with a mixture similar to Llama models~\cite{touvron2023llama}.
For evaluation, we consider a set of standard tasks: SciQ~\cite{welbl2017sciq}, TriviaQA~\cite{joshi2017triviaqa}, WebQ~\cite{berant-etal-2013-webqs}, MMLU~\cite{hendrycks2020mmlu}, GSM8k~\cite{cobbe2021gsm8k} LAMBADA~\cite{paperno2016lambada}, PiQA~\cite{bisk2020piqa}, HellaSwag~\cite{zellers2019hellaswag}, WinoGrande~\cite{sakaguchi2021winogrande}, ARC-easy (ARC-E) and ARC-challenge (ARC-C) \citep{arc-ce}.

\begin{wrapfigure}{l}{0.45\textwidth}  %
\vspace{-5mm}
\centering
\begin{tabular}{l|ccc}
\hline
\textbf{Parameters} & \textbf{3B} & \textbf{12B}  \\
\hline
d\_model & 2048 & 5120 \\
layers & 56 & 40 \\
num heads & 16 & 40 \\
num kv heads & 4 & 8  \\
qk-norm & yes & yes \\
head type & GQA & GQA  \\
head size & 128 & 128 \\
non-linearity & GeGLU & GeGLU  \\
feedforward dim & 6656 & 16384  \\
pre-norm & yes & yes \\
global-local ratio & 1:3 & 1:3 \\
\hline
\end{tabular}
\vspace{-2mm}
\caption{Model specifications of our 3B and 12B models.}
\vspace{-5mm}
\label{tab:model_spec}
\end{wrapfigure}
 
\paragraph{Model Setup} 
Our base model is a Transformer model with interleaved global-local attention layers.  The mixing ratio of global and local attention is always fixed at 1:3 with the first three layer being local and the final layer being global layer. The global layers do not use rotary positional embedding~\cite{su2024rope} whereas the local layers still use it with $\theta=5e5$. Such interleaved patterns are better in long context generalization based on our experience and literature~\cite{yang2025nope}. Apart from the positional embeddings, all the attention hyperparameters are shared between local and global attention.

In \sysname, all the attention hyperparameters for residual linear attention are shared from SWA (i.e., number of heads, number of kv heads, head\_dim) since query/key/value projections are directly  inherited from SWA for parameter efficiency. We emphasize again that \sysname does not require extra parameters compared with SWA or full attention.

\begin{figure}[t]
  \centering
  \begin{tikzpicture}
  \begin{axis}[
      width=0.8\textwidth,
      height=0.4\textwidth,
      xmode=log,
      log basis x={2},
      grid=both,
      grid style={line width=.1pt, draw=gray!10},
      major grid style={line width=.2pt,draw=gray!50},
      xlabel={Window Size},
      ylabel={MMLU Score (5-shot)},
      legend style={at={(0.02,0.98)}, anchor=north west},
      xtick={16,32,64,128,256,512,1024,2048},
      xticklabels={0,32,64,128,256,512,1024,2048},
      xmin=15, xmax=2300,
      ymin=0.32, ymax=0.45,
      legend cell align={left},
  ]
  \addplot[purple, dashed, line width=1.2pt] coordinates {
      (16, 0.368) (2048, 0.368)
  };
  \addlegendentry{\small{Global Attention Only}}
  \addplot[blue, mark=o, line width=1.2pt] coordinates {
      (2048, 0.369)
      (1024, 0.340)
      (512, 0.334)
      (256, 0.328)
  };
  \addlegendentry{\small{Local-Global \textit{w. SWA}}}
  \addplot[red, mark=o, line width=1.2pt] coordinates {
      (1024, 0.440)
      (512, 0.422)
      (256, 0.354)
      (128, 0.375)
      (64, 0.338)
      (32, 0.340)
      (16, 0.326)  %
  };
  \addlegendentry{\small{Local-Global \textit{w. RAttention}}}
  \addplot[brown, dashdotted, line width=1.2pt] coordinates {
      (16, 0.326) (2048, 0.326)
  };
  \addlegendentry{\small{Local-Global \textit{w. Linear Attention}}}
  \end{axis}
  \end{tikzpicture}
  \vspace{-2mm}
  \caption{Comparison of MMLU 5-shot performance scores across different window sizes at 3B scale with pretraining context length 4096. The horizontal purple dashed line represents the baseline using only global attention. The blue line shows Local-Global with sliding window attention (SWA), while the red line demonstrates the performance of Local-Global with \sysname. When window size $=0$, Local-Global  with \sysname reduces to Local-Global with only linear attention.}
  \vspace{-4.5mm}
  \label{fig:3b_perato_curve}
\end{figure}

\paragraph{Pareto Curve at 3B}

We train 3B-parameter SWA and \sysname models using various sliding window sizes on 400B tokens with a batch size of 1024. A full attention model is also trained as a baseline. As shown in Figure~\ref{fig:3b_perato_curve}, there is a clear tradeoff between window size and MMLU performance—smaller window sizes lead to reduced performance.

Importantly, \sysname models achieve a better tradeoff curve compared to SWA models. With a window size of $\geq 512$, \sysname already matches or even surpasses the performance of the full attention baseline. As we will show later, these gains persist across certain benchmarks when scaling up \sysname models. This observation is consistent with prior work suggesting that hybrid attention models can, in some cases, outperform standard Transformers~\cite{ren2025samba}.

\paragraph{Selective Pretraining at 3B and 12B}

We further verify the effectiveness of  \sysname  by scaling both the number of tokens, model parameters and context length.

\begin{wrapfigure}{l}{0.6\textwidth}  %
\vspace{-2mm}
\centering
\resizebox{0.6\textwidth}{!}{
\begin{tabular}{l|c|cc|c}
\textbf{Metric} & \textbf{Full 8k} & \textbf{SWA 4K} & \textbf{SWA 2K} & \textbf{RAttn-512}  \\
\hline
 \textbf{Average (0/1-shot)} & 62.67 & 62.64 & 62.18  & 62.72 \\
\hline
MMLU (5-shot) &  52.40 & 50.80 & 48.60 & 52.94 \\
 GSM8K (8-shot) & 36.69 & 35.71 & 33.28 & 37.39 \\
\hline
\end{tabular}%
}
\vspace{-2mm}
\caption{Main results at 12B scale with pretraining context length 8192. Performance of zero- and one-shot tasks are summarized in \textbf{Average (0/1-shot)}.}
\vspace{-3mm}
\label{tab:12b_8k_results}
\end{wrapfigure}
First, we selectively pretrain 3B-parameter SWA and \sysname models on 2T tokens.~\footnote{Our 400B/600B-token experiments takes 2-3 days to finish, and 2T-token experiments take 9 days.} As shown in Table~\ref{tab:3b_results}, \sysname with a window size of 512 outperforms SWA models using window sizes up to 2048. Second, we pretrain 12B-parameter SWA and \sysname models on 600B tokens and again evaluate performance across a range of window sizes. As shown in Table~\ref{tab:12b_results}, \sysname with a window size of 512 continues to outperform SWA models, further validating its scalability. Finally, we assess \sysname  with window size 512 in the setting of pretraining context length 8k and 600B tokens. The summary results shown in Table~\ref{tab:12b_8k_results} indicates that \sysname remains strong compared with full attention models.

\begin{table}[t!]
\centering
\resizebox{0.9\textwidth}{!}{%
\begin{tabular}{l|c|ccc|cc}
\textbf{Metric} & \textbf{Full 4k} & \textbf{SWA 2k} & \textbf{SWA 1k} & \textbf{SWA 512} & \textbf{RAtt 1k} & \textbf{RAtt 512} \\
\hline
ARC-C & 46.25 & 47.01 & 45.39 & 44.97 & 44.28 & 45.39 \\
ARC-E & 78.32 & 78.83 & 78.45 & 77.44 & 77.82 & 77.31 \\
HellaSwag & 56.61 & 56.02 & 56.10 & 56.21 & 56.43 & 56.43 \\
LAMBADA & 72.27 & 72.21 & 72.77 & 71.78 & 72.54 & 72.04 \\
PIQA & 78.56 & 78.24 & 78.56 & 78.89 & 78.45 & 78.24 \\
SciQ & 94.80 & 95.70 & 95.40 & 95.80 & 94.60 & 95.20 \\
WinoGrande & 68.98 & 68.82 & 68.67 & 71.03 & 71.27 & 70.80 \\
TriviaQA (1-shot) & 41.49 & 42.21 & 41.98 & 40.61 & 41.57 & 42.23 \\
WebQS (1-shot) & 20.37 & 19.34 & 20.37 & 21.70 & 18.31 & 18.55 \\
\rowcolor{gray!20} \textbf{Average (0/1-shot)} & 62.00 & 62.00 & 62.00 & 62.00 & 61.70 & 61.80 \\
\hline
\rowcolor{gray!20} MMLU (5-shot) & 56.70 & 55.70 & 55.70 & 55.50 & 56.22 & 55.62 \\
\rowcolor{gray!20} GSM8K (8-shot) & 34.19 & 29.49 & 32.90 & 27.98 & 33.13 & 33.74 \\
\hline
\end{tabular}%
}
\caption{Main results at 3B scale with pretraining context length 4096.}
\vspace{-4mm}
\label{tab:3b_results}
\end{table}
\begin{table}[t]
\centering
\resizebox{\textwidth}{!}{%
\begin{tabular}{l|c|c|ccccc}
\textbf{Metric} & \textbf{Full 4k} & \textbf{SWA 2K} & \textbf{RAttn-512} & \textbf{RAttn-256} & \textbf{RAttn-128} & \textbf{RAttn-64} & \textbf{RAttn-32} \\
\hline
ARC-C & 47.61 & 47.61 & 48.81 & 48.29 & 50.17 & 49.57 & 47.44 \\
ARC-E & 79.21 & 80.01 & 79.12 & 79.25 & 79.97 & 79.88 & 79.29 \\
HellaSwag & 57.76 & 57.99 & 57.72 & 57.79 & 58.11 & 58.25 & 58.06 \\
LAMBADA & 73.55 & 73.06 & 73.14 & 73.88 & 73.65 & 72.99 & 73.51 \\
PIQA & 79.27 & 80.03 & 78.51 & 79.54 & 79.00 & 79.38 & 79.43 \\
SciQ & 95.60 & 96.40 & 95.40 & 95.90 & 95.70 & 95.40 & 95.30 \\
WinoGrande & 70.17 & 72.69 & 71.03 & 71.74 & 72.38 & 71.90 & 70.09 \\
TriviaQA (1-shot) & 41.51 & 41.22 & 41.85 & 41.34 & 42.56 & 40.97 & 42.05 \\
WebQS (1-shot) & 21.21 & 21.65 & 24.66 & 22.64 & 22.00 & 20.42 & 20.23 \\
\rowcolor{gray!20} \textbf{Average (0/1-shot)} & 62.90 & 63.40 & 63.40 & 63.40 & 63.70 & 63.20 & 62.80 \\
\hline
\rowcolor{gray!20} MMLU (5-shot) &  52.96 & 49.52 & 51.50 & 53.77 & 51.74 & 49.17 & 49.66 \\
\rowcolor{gray!20} GSM8K (8-shot) & 30.33 & 24.26 & 29.57 & 29.34 & 26.61 & 30.93 & 26.61 \\
\hline
\end{tabular}%
}
\caption{Main results at 12B scale with pretraining context length 4096.}
\vspace{-7mm}
\label{tab:12b_results}
\end{table}

\subsection{Long-Context Results}

We then evaluate zero-shot generalization capability on the RULER~\cite{hsieh2024ruler} benchmark, testing them directly after pretraining at context length 4k. The average results are shown in Table~\ref{tab:ruler_results}.

\begin{wrapfigure}{l}{0.5\textwidth}
\vspace{-1mm}
\centering
\begin{tabular}{l|c|c|c|c}
\textbf{Model}& \textbf{4k} & \textbf{8k} & \textbf{16k} & \textbf{32k}  \\
\hline 
\textbf{Full-4k} & 80.38 & 2.89 & 0.00 & 0.08 \\
\textbf{SWA-2k} & 73.49 & 6.85 & 0.79 & 0.41 \\
\textbf{RAttn-1k} & 73.87 & 53.90 & 40.00 & 20.84  \\
\textbf{RAttn-512} & 80.79 & 66.26 & 50.80 & 29.59  \\
\end{tabular}
\label{tab:ruler_results}
\caption{Average zero-shot RULER performance at 3B scale with pretraining context length 4K.}
\vspace{-4mm}
\end{wrapfigure}
\sysname models generalize reasonably well beyond the 4k training context, whereas other models fail to do so. Interestingly, \sysname models with smaller window sizes exhibit better generalization. We believe that smaller window sizes place greater pressure on the local attention module to generalize beyond the local window during pretraining, resulting in improved length generalization.

\subsection{Training Efficiency}

We next demonstrate that the training efficiency of \sysname is not compromised. Specifically, we benchmark training efficiency in terms of step time using a batch size of 1024 and context lengths of 4k and 8k on TPU v5p-1024 (which is more suitable for larger-scale pretraining than v6e).
As shown in Table~\ref{tab:training_efficiency}, \sysname matches the training speed of both full attention and SWA models. Although \sysname introduces an additional RLA kernel to dispatch in the local attention layers compared to SWA, its small window size and the use of highly optimized RLA kernels collectively allow it to achieve comparable training speeds.

\begin{table}[htbp]
\centering
\begin{tabular}{c|c|lll}
\hline
\textbf{Model Size} & \textbf{Pretraining Length} & \textbf{Full Attention} & \textbf{SWA} & \textbf{\sysname} \\
\hline
\multirow{2}{*}{3B} & 4096 & 0.84 (4k) & 0.80 (2k) & 0.87 (512) \\
                    & 8192 & 1.20 (8k) & 1.05 (4k) & 1.08 (1k) \\
\hline
\multirow{2}{*}{12B} 
& 4096 & 2.21 (4k) & 2.10 (2k) & 2.26 (512) \\
                     & 8192 &  3.99 (8k) &  3.89 (4k) & 3.97 (1k) \\
\hline
\end{tabular}
\vspace{3mm}
\caption{Training speed comparison in terms of step time (seconds) at both 3B and 12B scale. Numbers in parentheses indicates window size of models.}
\vspace{-5mm}
\label{tab:training_efficiency}
\end{table}

\subsection{Inference Efficiency}

\begin{figure}[t]
\centering
\includegraphics[width=\textwidth]{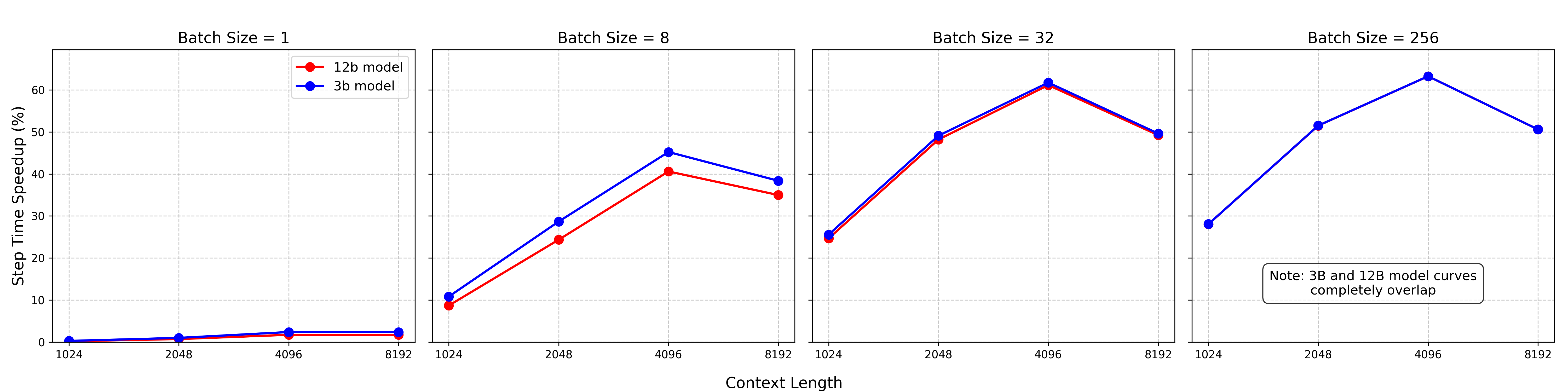}
\vspace{-5mm}
\caption{Step time speedup (\%) of local-global models using \sysname (window size 512) compared to SWA (window size 4k). As batch size increases, the theoretical speedup of \sysname increases and converges, since the KV cache size increasingly dominates the memory cost relative to the model parameter size.}
\label{fig:step_time}
\vspace{-3mm}
\end{figure}

Next, we analyze the inference gains achievable by \sysname when using a smaller sliding window size. Since the prefilling stage has a similar efficiency profile to the training stage, we focus our analysis on the step time during the generation stage. In this phase, the attention modules are typically memory-bound, while the feedforward modules can be either compute-bound or memory-bound depending on the batch size.
In general, the theoretical step time can be approximated by:
$$T_{\text{step}} = \frac{B \times S_{\text{KV}}}{BW} + \max\left(\frac{2 \times B \times P_{\text{count}}}{F}, \frac{P_{\text{size}}}{BW}\right)$$
where $T_{\text{step}}$ is the theoretical step time,  
$B$ is the batch size,  
$S_{\text{KV}}$ is the KV cache size,  
$BW$ is the total memory bandwidth,  
$P_{\text{count}}$ is the parameter count,  
$P_{\text{size}}$ is the parameter size (in bytes), and  
$F$ is the total FLOPs per second.
As a case study, we apply this analysis to our 3B and 12B models using H100 hardware specifications and bfloat16 precision. Figure~\ref{fig:step_time} shows the step time speedup as a function of context length across different batch sizes. As the batch size increases, the theoretical speedup of \sysname grows, reaching up to approximately 60\%.  Moreover, the speedup ultimately converges to the same point regardless of model size, since the KV cache size increasingly dominates the memory cost relative to the model parameter size.

\subsection{Ablation Study}

\begin{table}[t]
\centering
\label{tab:3b-results}
\resizebox{1.0\textwidth}{!}{%
\begin{tabular}{c|c|ccccccc}
\textbf{Metric} & \textbf{Full-4k} & \textbf{RAttn-512} & \textit{w.} ReLU & \textit{w.} Identity  & \textit{w.} Mamba2  & \textit{w.} Hymba & -SWA  & -GroupNorm \\
\hline
\textbf{Average (0/1-shot)} & 68.40 & 68.20 & 68.30 & 67.30 & 68.00 & 68.10 & 68.20 & 68.10 \\
\hline
\textbf{MMLU (5-shot)} & 36.80 & \textbf{42.22} & 37.90 & 35.70 & 39.21 & 36.53 & 32.60 & 39.81  \\
\hline
\end{tabular}%
}
\vspace{3mm}
\caption{Ablation study on 3B models with 400B tokens and context length 4096.}
\vspace{-8mm}
\label{tab:ablation}
\end{table}
We conducted ablation studies during 3B model training on the 400B token setting to identify the best configuration for \sysname. Results are shown in Table~\ref{tab:ablation}. For the feature map choice, we adopt $\softmax$ following \cite{zhang2025lolcats}, which we found to outperform other alternatives such as ReLU and Identity.

We also experimented with adding more complex gating mechanism to linear attention, specifically using Mamba2~\cite{dao2024mamba2} and Gated DeltaNet~\cite{yang2024gated}. However, no gain (or slight performance drop) is observed. We suspect that the introduction of more advanced linear models brings up optimization challenges: in our setup, the hybrid model already incorporates three token-mixing modules -- full attention, sliding window attention, and residual linear attention -- with the latter two sharing parameters. Introducing more complex forms of linear attention appears harder to optimize in this hybrid framework. On the other hand, we believe that it opens up a practical direction to explore for further research: designing better parameter-efficient linear models in the hybrid frameworks.

Additionally, we attempted to stack linear attention and sliding window attention across different heads, following the approach of Hymba~\cite{dong2024hymba}. However, this configuration proved suboptimal, likely because both mechanisms exhibit strong recency bias toward tokens within the sliding window. We also verified that linear attention alone cannot retrain the performance.
Finally, we found that applying group normalization improves the overall performance.

\subsection{Related Work}

Overall, there are two main approaches in designing efficient language models. The first relies on constant-memory modules, while the second focuses on leveraging sparsity in attention computation. %

\paragraph{Constant-Memory Models and Their Hybrids}
Recurrent models and SWA models serve as the primary building blocks for hybrid architectures due to their constant-memory properties. However, pure constant-memory models often underperform standard Transformers~\cite{vaswani2023xmr}. Early hybrid models integrated these modules by interleaving them with standard attention. For example, Gemma2/Gemma3~\cite{gemmateam2024gemma2,gemmateam2025gemma3} alternate SWA with global attention, while Jamba~\cite{lieber2024jamba} and Samba~\cite{ren2025samba} combine Mamba with either SWA or global attention. Similarly, Griffin~\cite{de2024griffin} integrates gated linear recurrences with SWA. Another line of research seeks to fuse attention and recurrent mechanisms within the same layer. Megalodon~\cite{ma2024megalodon} uses a recurrent model to refine query/key representations within attention, while Hymba~\cite{dong2024hymba} runs both Mamba and attention in parallel within each layer. Our approach, \sysname, advances this direction by achieving better parameter efficiency than Hymba—completely sharing parameters between linear attention and SWA. %

\paragraph{Sparse Attention Models}
To improve efficiency in long-context settings, another approach focuses on sparse attention, where the core challenge is designing effective sparsity patterns for KV-cache access. Early methods~\cite{roy2020routing,kitaev2020reformer,bertsch2023unlimiformer} use non-parametric techniques (e.g., k-nearest neighbors, locality-sensitive hashing) to select relevant query-key pairs. More recently, parametric methods that learn sparsity patterns have proven effective, such as Native Sparse Attention~\cite{yuan2025nsa} and Mixture of Block Attention (MoBA)~\cite{lu2025moba}. With hardware-aligned implementations, these modules can be trained more efficiently than global attention. However, unlike linear attention and SWA, sparse attention still requires storing the KV cache for all context tokens, same as global attention. This raises an ongoing research question: Is it more effective to sparsely access context within attention (as in sparse attention) or to rely on recurrent modules for context compression? We leave the investigation of this question for future work.
\section{Conclusion and Future Work}

In this work, we explore using \sysname to replace sliding window attention in local-global models. Our results show that residual linear attention enables a substantial reduction in sliding window size—from 4K/8K to 512—without loss in performance. Through both analytical and empirical studies on training and inference efficiency, we demonstrate that \sysname offers significant advantage over SWA: training efficiency is maintained, while inference efficiency is significantly improved.
In future work, we plan to focus on engineering efforts to realize the theoretical efficiency gains within current inference frameworks.
Fineuning existing pretrained full attention models into \sysname models is also promising direction to explore.

\section*{Acknowledgment}

We thank Sam Wiseman, Tao Lei, Aonan Zhang, Jianyu Wang, Karen Yang for their valuable feedback.

\bibliography{main}
\bibliographystyle{abbrv}

\end{document}